\documentclass[conference]{IEEEtran}
\IEEEoverridecommandlockouts

\usepackage{amsmath}
\usepackage{cite}
\usepackage{amsmath,amssymb,amsfonts}
\usepackage{algorithmic}
\usepackage{graphicx}
\usepackage{textcomp}
\usepackage{xcolor}
\usepackage{threeparttable}
\usepackage{booktabs}
\usepackage{multirow}
\usepackage{makecell}
\usepackage[table]{xcolor}
\usepackage{booktabs}
\usepackage{threeparttable}
\usepackage[table]{xcolor}
\usepackage{siunitx}
\usepackage{makecell}
\usepackage{array}
\usepackage{colortbl} 
\usepackage{siunitx}     
\usepackage[table]{xcolor} 
\usepackage{graphicx}
\usepackage{subcaption}
\usepackage{caption}
\usepackage{multirow}    
\usepackage{adjustbox}   
\usepackage{booktabs}    
\usepackage{amsmath}     
\usepackage{array}       
\usepackage[most]{tcolorbox}
\usepackage{amsmath}
\tcbuselibrary{breakable}
\usepackage{enumitem}
\newcommand{\OA}{\mathrm{OA}}
\newcommand{\mIoU}{\mathrm{mIoU}}

\usepackage{algorithm}
\usepackage{algorithmic}

\def\BibTeX{{\rm B\kern-.05em{\sc i\kern-.025em b}\kern-.08em
    T\kern-.1667em\lower.7ex\hbox{E}\kern-.125emX}}

\newcommand{\corrauthor}{\textsuperscript{$^{*}$}}

\def\BibTeX{{\rm B\kern-.05em{\sc i\kern-.025em b}\kern-.08em
    T\kern-.1667em\lower.7ex\hbox{E}\kern-.125emX}}
    
\begin{document}

\title{SOFTooth: Semantics-Enhanced Order-Aware Fusion for Tooth Instance Segmentation%
\thanks{\noindent\parbox{\linewidth}{%
\corrauthor~Corresponding author.}}%
}

\author{%
  \IEEEauthorblockN{%
    Xiaolan Li\textsuperscript{1},
    Wanquan Liu\textsuperscript{1},
    Pengcheng Li\textsuperscript{2},
    Pengyu Jie\textsuperscript{1} and Chenqiang Gao\textsuperscript{1}\corrauthor}
  \IEEEauthorblockA{%
    \textsuperscript{1}School of Intelligent Engineering, Sun Yat-sen University, China\\
    \textsuperscript{2}School of Computer Science and Technology, Hainan University, China}
}

\maketitle
\begin{abstract}
Three-dimensional (3D) tooth instance segmentation remains challenging due to crowded arches, ambiguous tooth-gingiva boundaries, missing teeth, and rare yet clinically important third molars. Native 3D methods relying on geometric cues often suffer from boundary leakage, center drift, and inconsistent tooth identities, especially for minority classes and complex anatomies. Meanwhile, 2D foundation models such as the Segment Anything Model (SAM) provide strong boundary-aware semantics, but directly applying them in 3D is impractical in clinical workflows. To address these issues, we propose SOFTooth, a semantics-enhanced, order-aware 2D--3D fusion framework that leverages frozen 2D semantics without explicit 2D mask supervision.
First, a point-wise residual gating module injects occlusal-view SAM embeddings into 3D point features to refine tooth-gingiva and inter-tooth boundaries.
Second, a center-guided mask refinement regularizes consistency between instance masks and geometric centroids, reducing center drift.
Furthermore, an order-aware Hungarian matching strategy integrates anatomical tooth order and center distance into similarity-based assignment, ensuring coherent labeling even under missing or crowded dentitions. 
On 3DTeethSeg’22, SOFTooth achieves state-of-the-art overall accuracy and mean IoU, with clear gains on cases involving third molars, demonstrating that rich 2D semantics can be effectively transferred to 3D tooth instance segmentation without 2D fine-tuning.
\end{abstract}

\begin{IEEEkeywords}
Tooth segmentation, 2D--3D fusion, instance segmentation, third molars
\end{IEEEkeywords}

\section{Introduction}

Computer-aided design (CAD) technology has become integral to modern digital dentistry, providing precise support for critical clinical procedures such as orthodontic treatment and dental implantation~\cite{1,CADtiaozhan}. As a foundational prerequisite for these CAD-driven treatment processes, three-dimensional (3D) tooth instance segmentation enables automated diagnosis, personalized treatment planning, and prognosis evaluation in digital dentistry~\cite{qianti,2dcnn,darch,li2023thisnet}. Optical intraoral scanners (IOS), which capture high-fidelity 3D geometric information of teeth and gingiva without radiation, are now widely integrated into dental CAD systems, enabling efficient and routine generation of digital dental models~\cite{IOSyingyong}. 
However, achieving high-precision automatic segmentation remains difficult due to the inherent complexities of dental anatomy. Teeth are typically arranged in tight, curved arches, resulting in ambiguous inter-tooth boundaries that are difficult to distinguish. Moreover, the demand for fine-grained segmentation to satisfy clinical standards further exacerbates this difficulty, as even subtle boundary inaccuracies may compromise subsequent treatment planning. These factors make accurate segmentation challenging in real-world clinical scenarios.

\begin{figure}
    \centering
    \includegraphics[width=\linewidth]{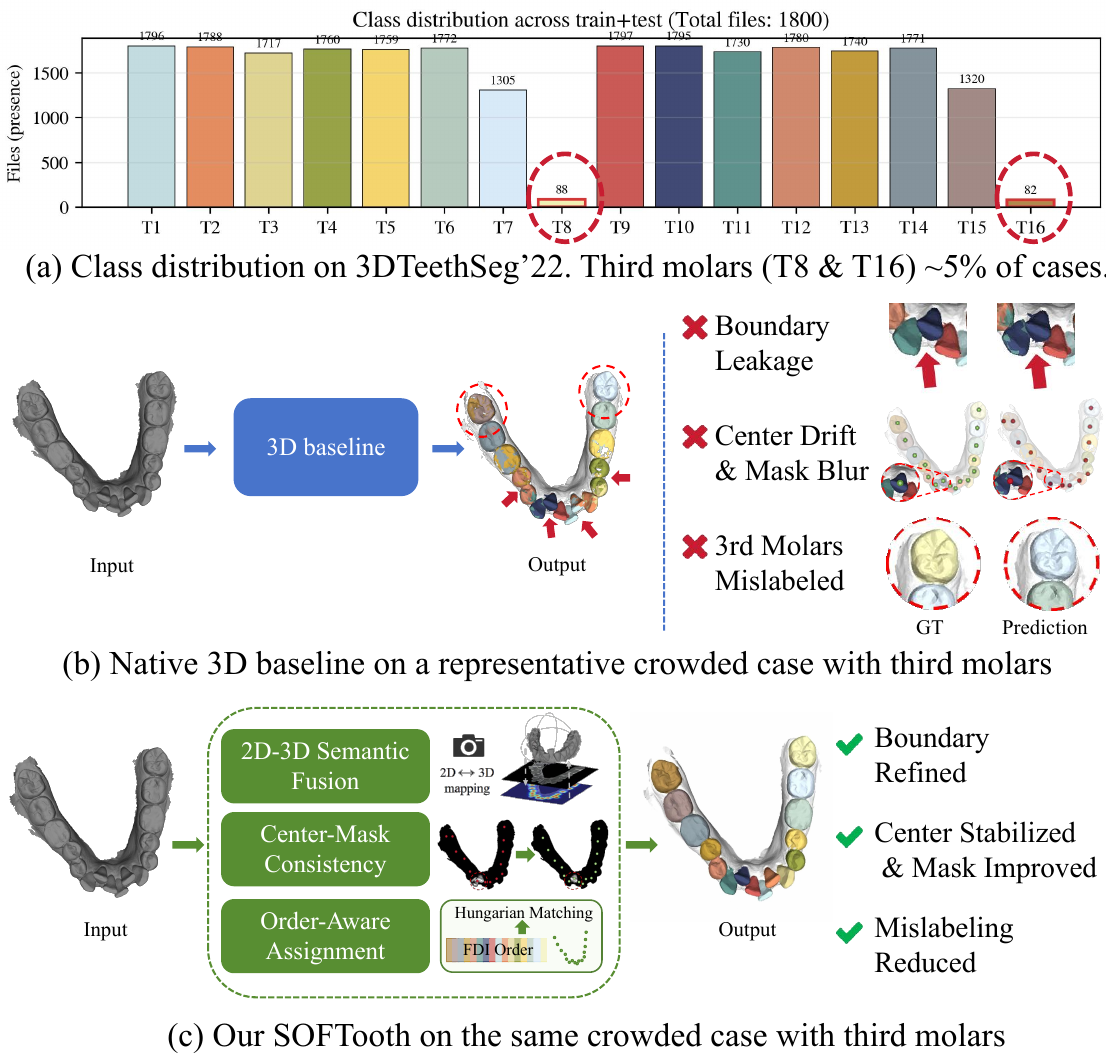}
    \caption{Existing limitations vs. our improvements.
(a) The benchmark dataset 3DTeethSeg'22 shows an imbalanced class distribution, with third molars appearing in only about 5\% of cases.
(b) Native 3D baselines predominantly rely on 3D geometric features (e.g., coordinates, normals) without effective semantic guidance or structural constraints, failing to address the spatial complexity of crowded 16-tooth cases.
(c) Our method exploits 2D--3D semantic fusion for boundary refinement, center-mask consistency to stabilize centers and improve mask reliability, and order-aware assignment to maintain anatomical tooth order and reduce mislabeling of third molars.}

    \label{fig:intro}
\end{figure}
Recent efforts to address these challenges have primarily focused on native 3D methods that operate directly on points, faces, or meshes and employ attention mechanisms to preserve geometric structure and local curvature or normal information~\cite{TSGC,dbganet,tsrnet,teethgnn}.
While these methods have demonstrated strong performance on regular dental cases, their efficacy degrades substantially in clinically challenging scenarios, particularly when handling rare yet clinically critical third molars and other complex anatomical variations.

Third molars, a representative minority class in tooth segmentation, are clinically important for treatment decisions such as extraction planning and impaction assessment. Yet they appear in only a small fraction of clinical scans, leading to class imbalance, as illustrated by the 3DTeethSeg'22 benchmark dataset~\cite{3dteethseg}, where cases containing third molars account for only about 5\% of all scans (Fig.~\ref{fig:intro}(a)).
Owing to their morphological similarity to adjacent second molars, native 3D methods, which predominantly rely on 3D geometric features (e.g., coordinates, normals), struggle to capture the fine-grained discriminative cues required to distinguish them. As a result, third molars are frequently misclassified as second molars, yielding low recall for this clinically important minority class~\cite{2018Imbalance,2019ImbalanceSurvey}. Furthermore, clinical anomalies such as missing teeth, dental crowding, and incomplete eruption often co-occur with third molars and yield irregular dentitions. These complex cases require models to capture fine-grained semantic cues and leverage global structural and anatomical priors. However, existing native 3D methods do not adequately satisfy these requirements: they (i) exhibit low recall for visually similar teeth and minority classes, often accompanied by boundary leakage in crowded regions, (ii) suffer from center drift and blurred boundaries, and (iii) produce inconsistent tooth identities due to the absence of explicit spatial or anatomical ordering (Fig.~\ref{fig:intro}(b)).

To address these challenges, we propose SOFTooth, a unified framework that integrates boundary-sensitive 2D semantics, stable instance centers, and anatomical order into a single end-to-end pipeline (Fig.~\ref{fig:intro}(c)). This design enables robust 3D tooth instance segmentation for both normal cases and complex cases with missing teeth, crowding, or third molars. 
Specifically, motivated by recent works showing that 2D foundation models such as SAM provide strong boundary-sensitive cues and that image-derived semantics can enhance 3D boundary delineation~\cite{iossam,SAMtooth,medsam}, we propose three complementary components.
First, a Point-wise Residual Gating (PRG) module renders an occlusal view, samples frozen SAM embeddings, and injects them into native 3D features as boundary-sensitive priors. A geometry-aligned projection together with center-guided, instance-aware gating further aligns these 2D semantics with the 3D geometry and concentrates them on a single tooth region, which enhances contrast along tooth-gingiva and inter-tooth boundaries and alleviates instance leakage in crowded arches.
Second, based on the PRG-enhanced representation, our Center-Guided Mask Refinement (CMR) module decodes instance queries with dynamic mask kernels, derives 3D instance centers from the predicted masks, and enforces center-mask geometric consistency via class-consistent pseudo-mask supervision and a center-based regularizer, yielding stable instance centers and clean masks in crowded or partially erupted regions.
Third, given the refined centers and masks from CMR, we propose an FDI Order-Aware Hungarian Matching (FHM) module that incorporates FDI tooth sequence and center distance into the instance matching cost, encouraging assignments that are consistent with anatomical tooth order and providing anatomically coherent supervision.
By coupling PRG-enhanced features, center-consistent masks, and order-aware supervision, SOFTooth can reliably separate rare third molars from neighboring second molars and maintain consistent identities even in scans with missing teeth, crowding, or incomplete eruptions.

The main contributions of our method can be summarized as follows:
\begin{itemize}
\item We propose SOFTooth, a dual-branch 2D--3D fusion framework that integrates 2D semantics, geometric centers, and anatomical tooth order into a single end-to-end pipeline for 3D tooth instance segmentation.
\item We propose three complementary components: PRG for injecting frozen SAM occlusal-view semantics into 3D features, CMR for improving center-mask consistency, and FHM for anatomical order–aware instance assignment. They jointly reduce boundary leakage, center drift, and identity confusion in both regular cases and challenging cases with missing teeth, crowding, and third molars.
\item Extensive experiments demonstrate that SOFTooth achieves state-of-the-art performance, significantly improving recall for minority classes and maintaining robust performance in challenging scenarios, such as incomplete or heavily crowded dentitions.
\end{itemize}

\section{Related Work}

\subsection{3D Tooth Semantic Segmentation and Cross-Modal Fusion}
Conventional methods for 3D dental model segmentation mainly rely on handcrafted geometric priors, such as morphological skeletons~\cite{wu2014tooth}, harmonic fields~\cite{zou2015}, and curvature-driven region growing~\cite{li2007fast}, which delineate individual teeth and gingiva~\cite{2008orthodontics,RAITH201765,kumar2011improved}. These methods define explicit rules over pre-selected geometric features (e.g., coordinates, normals, curvature) and often require manual landmark annotation or human--computer interaction, which limits their robustness on challenging 3D dental models~\cite{dbganet}.

Recent advances in 3D deep learning have led to methods that operate directly on raw 3D point clouds or meshes for tooth segmentation. Point-based approaches build on general architectures such as PointNet~\cite{pointnet} and subsequent point-based variants~\cite{pointnet++,votenet,jsnet,lcpformer}, adapting these frameworks to dental models for point-wise feature learning and tooth-gingiva classification.
Mesh-based networks~\cite{meshseg,imeshseg,meshlab,surface,3DShape} perform semantic segmentation on cells or vertices to capture multi-scale local context along surfaces and are readily applicable to tooth meshes. 
In this line of work, TSGCNet~\cite{TSGC} pioneers a two-stream graph convolutional architecture that learns coordinate and normal features in separate branches, enriching geometric descriptors and improving tooth-gingiva boundary discrimination. 
Subsequent graph-based models~\cite{dbganet,tsrnet} further refine geometric encoding with multi-scale aggregation and dual-branch geometric attention, yielding richer 3D representations and sharper boundaries on dental models. 
These native 3D semantic methods substantially improve boundary modeling and overall segmentation performance on intraoral scans. However, when teeth are tightly arranged or morphologically similar, purely geometric features still struggle to capture fine-grained boundary cues, leading to boundary leakage and mislabeling between adjacent teeth.

Beyond purely geometric modeling, a complementary line of work explores cross-modal fusion between images and 3D dental models. Early transformation-based pipelines (e.g., 2D projection~\cite{2dcnn} and voxelization~\cite{voxel}) improve network compatibility but can reduce geometric fidelity near tooth-gingiva interfaces and tight contacts. 
With recent advances in 2D foundation models, the Segment Anything Model (SAM)~\cite{sam} has been proposed for promptable, category-agnostic segmentation. Recent cross-modal studies~\cite{iossam,SAMtooth,2025cvpr} have shown that transferring image-derived semantics back to 3D can enhance boundary delineation: IOSSAM~\cite{iossam} uses SAM as a 2D boundary prior and lifts multi-view masks to 3D via graph diffusion, while CrossTooth~\cite{2025cvpr} fuses features from rendered images back onto points and reports higher boundary IoU. These methods demonstrate that image semantics can effectively complement pure geometric features for boundary refinement, but they often involve multi-view rendering and dedicated 2D decoders.

Our approach is related to these native 3D and cross-modal pipelines but differs in fusion strategy. The proposed PRG module injects frozen SAM occlusal-view embeddings as boundary-sensitive, instance-aware priors into native 3D point features via geometry-aligned, point-wise residual gating, refining tooth-gingiva and inter-tooth boundaries and reducing instance leakage across adjacent teeth, without any SAM fine-tuning or tooth-specific 2D mask supervision.

\subsection{3D Tooth Instance Segmentation}   
Beyond 3D tooth semantic segmentation, another important direction is 3D tooth instance segmentation. 
Center-based methods~\cite{Mask-MCNet,teethgnn,cui2021tsegnet,darch} predict tooth centers, offset fields, or 3D proposals and then group points according to offset-to-center consistency or proposal assignment. 
For example, TSegNet~\cite{cui2021tsegnet} combines distance-aware centroid voting with cascaded centroid regression and confidence refinement to obtain robust single-tooth masks on point clouds, while DArch~\cite{darch} adopts a two-stage architecture that leverages dental-arch priors to first detect tooth centroids along the arch and then refine instance masks guided by this structural constraint. 
More recently, affinity-based methods such as THISNet~\cite{li2023thisnet} leverage tooth object affinity maps together with global contextual cues to highlight regions corresponding to each tooth, enabling end-to-end tooth instance segmentation and labeling on 3D dental models.
These instance-based approaches effectively exploit geometric priors and contextual cues to improve separation between contacting teeth and achieve strong performance when boundaries are clear and class distributions are relatively balanced. 

However, under complex clinical conditions with missing or crowded teeth and rare third molars, the association between predicted centers and masks often becomes unstable: centers may drift away from true tooth centroids, and instance masks may no longer be tightly aligned with their centers. This center-mask inconsistency can propagate errors to subsequent grouping and matching, causing fragmented instances and misassigned identities in crowded or partially erupted regions.
To address these problems, we propose a CMR module that decodes a fixed set of instance queries with dynamic mask kernels, derives 3D centers from the predicted masks, and jointly refines mask logits and centers under pseudo-mask supervision and a center-consistency regularizer. By explicitly enforcing center-mask consistency, the CMR module stabilizes instance centers and yields cleaner masks, particularly in challenging cases with crowded dentitions or incomplete eruptions.

\begin{figure*}[tb]
    \centering
    \includegraphics[width=\textwidth]{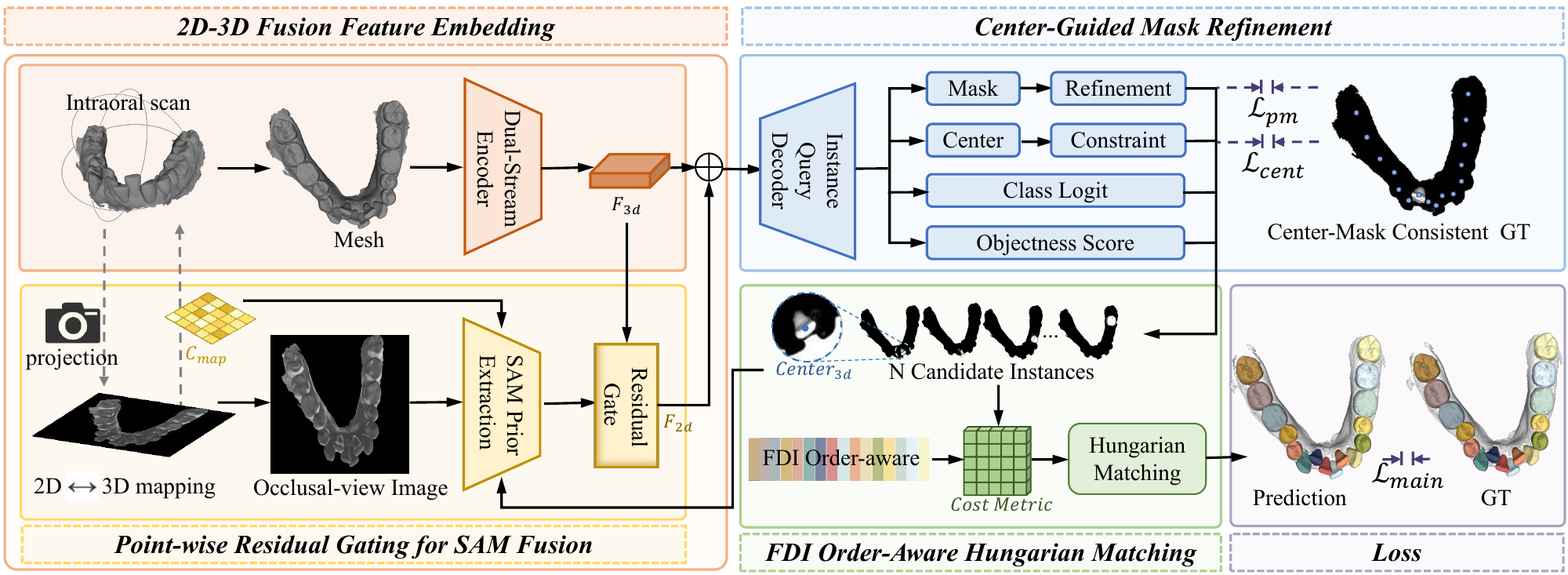}
    \caption{Architecture of SOFTooth for 3D tooth instance segmentation. A dual-stream 3D backbone extracts geometric features from intraoral meshes. The Point-wise Residual Gating (PRG) module renders an occlusal view, samples frozen SAM embeddings via geometry-aligned projection, and injects them into the 3D features through center-guided, point-wise residual gating, yielding boundary-sensitive fused representations that sharpen inter-tooth and tooth-gingiva boundaries. The Center-Guided Mask Refinement (CMR) module learns instance queries and dynamic kernels, and aligns masks with 3D geometric centers to correct center drift and refine tooth instances. The FDI Order-Aware Hungarian Matching (FHM) module performs order-aware matching in similarity space with integrated center distance, yielding anatomically coherent tooth indices for supervision.}

    \label{fig:placeholder}
\end{figure*} 

\subsection{Anatomical Priors and FDI-Aware Labeling}
Beyond segmentation quality, anatomical priors and the two-digit notation of the FDI World Dental Federation (F\'ed\'eration Dentaire Internationale; FDI notation)~\cite{ISO3950} are crucial for reliable tooth-identity assignment in clinical workflows.
On 2D radiographs, several methods~\cite{tekin2022enhanced},~\cite{chen2019deep} combine CNN-based tooth detection or instance segmentation with FDI-based post-processing rules to assign tooth numbers. 
For 3D intraoral scans, some approaches~\cite{srf,tseglab,yagong} exploit pairwise tooth-offset cues or spatial relationship features to enforce FDI-consistent sequences for labeling. 

Collectively, these methods exploit the FDI system as a structural constraint to guide tooth identification, but FDI priors are mainly imposed at the label level as post-hoc corrections after mask prediction, rather than being integrated into the instance matching or feature learning process itself.
In contrast, our FHM module incorporates FDI indices and center distances directly into the instance matching cost, providing anatomy-aware supervision during training rather than only at post-processing. By jointly considering spatial proximity and anatomical order during assignment, FHM prevents rare third molars from being misclassified as adjacent second molars and improves identity stability under missing or irregular dentitions, complementing the refined centers and masks produced by our CMR module.

\section{Method}
\label{sec:method}
\subsection{Overview}
Fig.~\ref{fig:placeholder} illustrates the architecture of the proposed network. First, we utilize a dual-stream dynamic graph convolutional geometry encoder to extract features from the 3D dental model. Then, we propose the PRG module for 2D--3D fusion, which samples frozen SAM embeddings at per-face pixel locations and fuses them into 3D point features through a gated residual mechanism. The CMR module then refines instance masks and stabilizes center localization by enforcing geometric center consistency. Finally, our FHM module incorporates anatomical order constraints and center distances directly into the matching cost, providing anatomy-aware supervision during training rather than only at post-processing, enabling one-to-one matching and accurate labeling. Together, these components facilitate reliable and precise tooth instance segmentation and labeling on 3D dental models.

\subsection{2D--3D Fusion Feature Embedding}
The 2D--3D fusion feature embedding module integrates 3D geometric features with 2D semantic priors extracted from a frozen SAM encoder.
It consists of a dual-stream 3D feature encoder and a point-wise residual gating (PRG) block for 2D--3D fusion.

\subsubsection{Dual-Stream 3D Feature Extraction}
Inspired by TSGCNet~\cite{TSGC}, we employ a dual-stream dynamic graph convolutional geometry encoder to extract 3D features from the dental mesh. For each mesh face \( i \), we create a 24-dimensional feature vector \( \mathbf{p}_i^{(0)} \in \mathbb{R}^{24} \) by concatenating the 3D coordinates and normals of the face center and its three vertices. Stacking over all faces yields the input tensor \( \mathbf{P} \in \mathbb{R}^{24 \times M} \).

The coordinate and normal channels are processed by two parallel streams. Each stream is first aligned by a learnable feature transform module (FTM) that predicts a $k \times k$ feature transformation matrix for feature normalization. The transformed features are then passed through three stages of $k$-NN graph blocks with attention-based aggregation to capture multi-scale geometric context. Let $\mathbf{p}_i^{(l)}$ denote the feature of face (graph node) $i$ at layer $l$, and let $\mathcal{N}_k(i)$ be the set of its $k$ nearest neighbors; index $j$ enumerates neighbors $j \in \mathcal{N}_k(i)$. The local attentive aggregation is computed as
\begin{equation}
\begin{aligned}
\mathbf{e}_{ij} &= \phi\!\left([\mathbf{p}_i^{(l)} - \mathbf{p}_j^{(l)},\, \mathbf{p}_j^{(l)}]\right), \\
\alpha_{ij} &= \mathrm{softmax}_{j \in \mathcal{N}_k(i)}(\mathbf{e}_{ij}), \\
\mathbf{g}_i^{(l)} &= \sum_{j \in \mathcal{N}_k(i)} \alpha_{ij} \, \psi(\mathbf{p}_j^{(l)}),
\end{aligned}
\end{equation}
where \( \phi \) and \( \psi \) are shared \( 1 \times 1 \) convolutions. The coordinate and normal streams are fused channel-wise, followed by point-wise self-attention to produce the global point features \( \mathbf{{F}_{3d}} \in \mathbb{R}^{ 128 \times M} \), which are used as input to the following module. This encoder can be replaced by other point cloud backbones (PointNet~\cite{pointnet}/PointNet++~\cite{pointnet++}, etc.).

\subsubsection{Point-wise Residual Gating for 2D--3D Fusion}
\begin{figure}
    \centering
    \includegraphics[width=\linewidth]{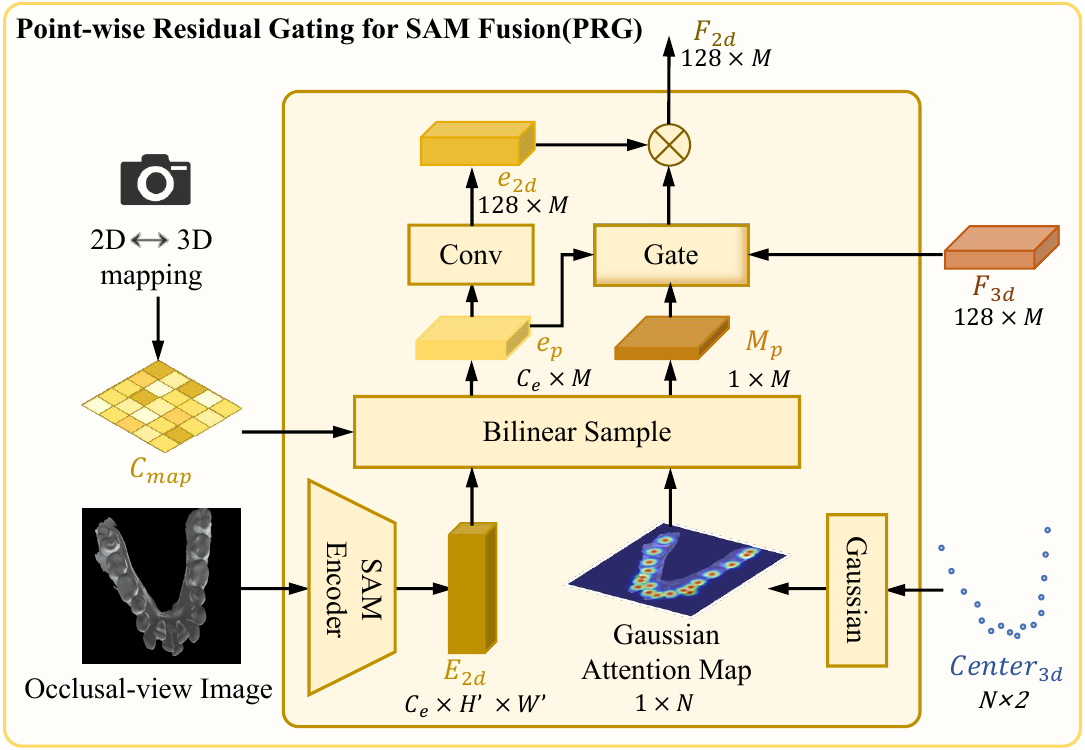}
    \caption{The structure of Point-wise Residual Gating for 2D--3D Fusion (PRG) module. Face projections on the occlusal view are used to bilinearly sample per-face SAM features and center-guided Gaussian weights, which serve as 2D priors on the mesh. A point-wise gating block then injects these priors into 3D face features via a residual connection, sharpening tooth-gingiva and inter-tooth boundaries.}
    \label{fig:prg}
\end{figure}

The PRG module effectively integrates the frozen SAM embeddings with the 3D point features in an instance-aware manner.
As shown in Fig. \ref{fig:prg}, given an intraoral mesh, the PRG module builds a one-to-one correspondence between each mesh face and a 2D occlusal-view pixel, samples SAM embeddings and fuses them into the 3D backbone via a gated residual mechanism. This design leverages occlusal-view cues to refine inter-tooth and tooth-gingiva boundaries and reduces instance leakage across adjacent teeth.
We first render an occlusal-view RGB image \(\mathbf{I} \in \mathbb{R}^{3 \times H \times W}\) of the intraoral scan, where the camera looks at the mesh centroid from above the occlusal plane to expose tooth–to–tooth and tooth–to–gingiva contacts. The rendered image is resized to match SAM’s input resolution and passed through the frozen SAM encoder, yielding a dense embedding \(\mathbf{E}_{2\mathrm{d}} \in \mathbb{R}^{C_e \times H' \times W'}\) with \(C_e = 256\).

After occlusal-view rendering, each mesh face $p$ has a corresponding pixel coordinate $(y_p, x_p)$ on the occlusal-view image of resolution $(H, W)$. Stacking all faces over the batch gives a coordinate map
\begin{equation}
\mathbf{C}_{\text{map}} = \{(y_p, x_p)\}_{p=1}^{M} \in \mathbb{R}^{M \times 2}.
\end{equation}
Since $(H', W')$ is typically lower than $(H, W)$, we first linearly rescale the pixel coordinates to the embedding grid:
\begin{equation}
\tilde{y}_p = y_p \cdot \frac{H' - 1}{H - 1}, 
\qquad
\tilde{x}_p = x_p \cdot \frac{W' - 1}{W - 1}.
\end{equation}

The continuous coordinates are then normalized and used to bilinearly sample $\mathbf{E_{2d}}$, obtaining per-face SAM features $\mathbf{e_p} \in \mathbb{R}^{ C_e \times M}$ that provide a dense semantic prior on the 3D surface. This sampling ensures reliable 2D--3D alignment while keeping the 2D prior frozen and label-free.

To provide instance-aware semantic priors for both 2D--3D fusion and mask refinement, we convert the predicted instance centers \( \{(y_k, x_k)\}_{k=1}^{K} \) on the occlusal view into soft spatial guidance maps. These maps highlight the areas around predicted instance centers. We then sample these maps at the face centers using the coordinate map \( \mathbf{C}_{\text{map}} \), obtaining per-instance, per-face weights \( w_{k,p} \in [0,1] \). Aggregating over instances produces a scalar guidance weight:

\begin{equation}
M_p = \mathcal{A}\big(\{w_{k,p}\}_{k=1}^{K}\big) \in \mathbb{R}^{ 1 \times M},
\end{equation}
where \( \mathcal{A} \) is the aggregation function that combines the per-instance, per-face weights into a single scalar value \( M_p \). This weight \( M_p \) highlights the faces that are spatially close to any predicted instance center on the occlusal view.

To inject 2D semantics into the 3D point features, we use a residual-gated injection mechanism:
\begin{equation}
\mathbf{F_{2d}}
=
\tau \, g\!\big([\mathbf{F_{3d}} \,\Vert\, \mathbf{e_p} \,\Vert\, M_p]\big)
\odot
\mathbf{e_{3d}},
\label{eq:residual_gated_injection}
\end{equation}
\begin{equation}
\mathbf{F_{\text{fused}}} = \mathbf{F_{3d}} + \mathbf{F_{2d}},
\end{equation}
where \( \mathbf{F}_{3d} \), \( \mathbf{F}_{2d} \), \( \mathbf{F}_{\text{fused}} \), and \( \mathbf{e_{3d}} \) are all in \( \mathbb{R}^{128 \times M} \), and \( \mathbf{e_{3d}} \) is the transformed version of the SAM embedding \( \mathbf{e_p} \). The gating network \( g(\cdot) \) is a two-layer convolutional block that outputs a per-face gating coefficient, and \( \tau \) is a scalar factor that modulates the overall strength of SAM integration. The element-wise multiplication (\( \odot \)) ensures that the contribution of the SAM-derived feature is spatially adaptive.

The fused features \( \mathbf{F}_{\text{fused}} \) combine 3D geometric and 2D semantic information and are fed into the subsequent instance module for mask refinement and center prediction, enabling effective 2D--3D fusion without relying on any tooth-specific 2D annotations. 
The PRG module encourages SAM semantics to be injected in a way that is both boundary-sensitive and instance-consistent, which in practice reduces instance leakage across adjacent teeth and yields sharper inter-tooth and tooth-gingiva boundaries in the fused features.

\subsection{Center-Guided Mask Refinement}
The CMR module improves instance masks and center localization by enforcing geometric center consistency. Given the fused features \( \mathbf{F}_{\text{fused}} \), we generate instance-level queries from the attention maps predicted by the network. The attention maps are computed as follows:
\begin{equation}
\mathbf{A} = \sigma\big(\mathrm{Conv}_{1\times 1}(\mathbf{F}_{\text{fused}})\big) \in \mathbb{R}^{K \times M},
\end{equation}
where \(K{=}120\) is the number of instance slots, \(\mathrm{Conv}_{1\times 1}(\cdot)\) is a shared point-wise convolution, and \(\sigma(\cdot)\) denotes a sigmoid that produces attention weights in \([0,1]\) for each face and slot. After normalizing \(\mathbf{A}\) along the face dimension, we aggregate instance descriptors by attention-weighted pooling:
\begin{equation}
\mathbf{F}_{\text{inst}} = \mathbf{A} \, \mathbf{F}_{\text{fused}}^{\top} \in \mathbb{R}^{ K \times 128},
\end{equation}
where \(\mathbf{F}_{\text{fused}}^{\top} \in \mathbb{R}^{M \times 128}\) is obtained by transposing the last two dimensions of \(\mathbf{F}_{\text{fused}}\). Each row of \(\mathbf{F}_{\text{inst}}\) summarizes a subset of faces as an instance-level query. These \(K\) queries are then passed through an instance decoder with a bottleneck that reduces the slots from \(K{=}120\) to \(N{=}30\) queries. The decoder attaches three prediction heads to each query, producing dynamic mask kernels, class logits, and an objectness score. In addition, a geometry-based center estimation branch derives 3D instance centers from the predicted masks and mesh coordinates. Together, these outputs define \(N\) candidate instances.

The dynamic mask kernels are applied to the projected mask features to generate per-instance mask logits
\(\mathbf{P} \in \mathbb{R}^{N \times M}\). For each instance, we binarize its mask logits to obtain the activated faces and compute the 3D centroid from their coordinates. The face whose 3D position is closest to this centroid is selected as the refined instance center, and its corresponding 2D coordinates are retrieved from the per-face occlusal-view projections. This procedure enforces mask--center geometric consistency, effectively correcting center drift and providing reliable instance centers for subsequent refinement stages.

To further stabilize the masks and centers, we introduce a class-consistent pseudo-mask supervision. We construct a pseudo-mask tensor \(\mathbf{M}_{\text{cls}} \in \{0,1\}^{N \times M}\) by assigning to each instance the face label of the point nearest to its predicted 2D center and expanding this label to a global binary mask. The pseudo-mask loss is defined as
\begin{equation}
\mathcal{L}_{\text{pm}} = \mathcal{L}_{\text{BCE}}(\mathbf{P}, \mathbf{M}_{\text{cls}}),
\end{equation}
where \(\mathcal{L}_{\text{BCE}}\) denotes the binary cross-entropy loss with logits.
In addition, we define a Smooth~L1 center regularization as:
\begin{equation}
\mathcal{L}_{\mathrm{cent}}
=
\big\|
\mathbf{c}^{3\mathrm{D}}_{\mathrm{match}}
-
\hat{\mathbf{c}}^{3\mathrm{D}}_{\mathrm{match}}
\big\|_1,
\label{eq:center_reg}
\end{equation}
where $\mathbf{c}^{3\mathrm{D}}_{\mathrm{match}}$ and $\hat{\mathbf{c}}^{3\mathrm{D}}_{\mathrm{match}}$ denote the derived 3D centers of matched predictions and the corresponding class-mean ground-truth centers, respectively.
It aligns the derived 3D centers with class-mean ground-truth centers during assignment, further encouraging consistent instance masks and stable centers.

\subsection{FDI Order-Aware Hungarian Matching}
We formulate prediction--target assignment in a similarity space that jointly considers mask quality, class confidence, 3D center proximity, and anatomical tooth order. 
Given \(N{=}30\) predicted instances and \(L\) ground-truth teeth, we first construct a base similarity matrix \(\mathbf{C} \in [0,1]^{N \times L}\) that combines mask overlap and classification confidence. 
Let \(\mathbf{P} \in \mathbb{R}^{N \times M}\) be the predicted mask logits, \(\mathbf{T} \in \{0,1\}^{L \times M}\) the ground-truth masks, and \(\mathbf{L} \in \mathbb{R}^{N \times C}\) the classification logits over \(C\) tooth classes.
We compute
\begin{equation}
\mathbf{C}
=
\big(S_{\mathrm{mask}}\big)^{\alpha}
\odot
\big(S_{\mathrm{cls}}\big)^{\beta},
\end{equation}
where \(S_{\mathrm{mask}} \in [0,1]^{N \times L}\) contains Dice scores between \(\sigma(\mathbf{P})\) and \(\mathbf{T}\), and \(S_{\mathrm{cls}} \in [0,1]^{N \times L}\) selects the predicted class probabilities \(\sigma(\mathbf{L})\) at the ground-truth label indices. 
The hyper-parameters \(\alpha\) and \(\beta\) control the balance between segmentation quality and classification confidence and are empirically set to \(0.8\) and \(0.2\), respectively.
This construction yields a similarity measure that favors predictions that are consistent in both mask shape and class label.

To encourage anatomically coherent assignments, we incorporate 3D center distances into the similarity. 
Let \(\mathbf{c}_i^{3\mathrm{D}} \in \mathbb{R}^3\) and \(\hat{\mathbf{c}}_l^{3\mathrm{D}} \in \mathbb{R}^3\) denote the derived 3D centers of predicted instance \(i\) and ground-truth tooth \(l\), respectively. 
With scene-scale normalization and a drift tolerance margin, we define a center-refined similarity:
\begin{equation}
\mathbf{C}_{\mathrm{cent}}(i,l)
=
\mathbf{C}(i,l)
-
\lambda_{\mathrm{cent}} \cdot
\rho\!\Bigg(
\frac{\|\mathbf{c}_i^{3\mathrm{D}} - \hat{\mathbf{c}}_l^{3\mathrm{D}}\|_1}{\mathcal{S}_{\text{scene}}}
-
\delta_{\mathrm{drift}}
\Bigg),
\label{eq:Ccenter}
\end{equation}
where \(\rho(x) = \max(x,0)\) denotes the positive-part (ReLU) function, 
\(\lambda_{\mathrm{cent}}\) controls the strength of the center penalty, 
\(\delta_{\mathrm{drift}}\) defines a tolerance margin for center deviations, 
and \(\mathcal{S}_{\text{scene}}\) is a scene-level normalization factor (e.g., the diagonal length of the dental model bounding box).
In all experiments, we set \(\lambda_{\mathrm{cent}}{=}0.5\) and \(\delta_{\mathrm{drift}}{=}0.10\).

We further incorporate an anatomical tooth-ordering prior derived from the FDI notation. 
Each scan is represented in a canonical dental coordinate frame in which the mesiodistal direction along the dental arch approximately aligns with the $x$-axis.
We therefore project the predicted 3D centers $\mathbf{c}_i^{3\mathrm{D}}$ onto this axis and sort them along the arch to obtain normalized ranks $\mathbf{r} \in [0,1]^{N}$. 
For the ground truth, FDI tooth indices are mapped to canonical ordinal positions along the arch (from central incisor to third molar within each quadrant) and normalized to $\mathbf{t} \in [0,1]^{L}$. 
The discrepancy between $\mathbf{r}_i$ and $\mathbf{t}_l$ measures how inconsistent a prediction--target pair is with the expected anatomical order, and we penalize large discrepancies via a piecewise-linear term subtracted in the similarity space:
\begin{equation}
\mathbf{C}_{\mathrm{ord}}(i,l)
=
\mathbf{C}_{\mathrm{cent}}(i,l)
-
\lambda_{\mathrm{ord}}\,
\rho\!\big(|\mathbf{r}_i - \mathbf{t}_l| - \delta_{\mathrm{ord}}\big),
\label{eq:Cord}
\end{equation}
where \(\lambda_{\mathrm{ord}}\) weights the FDI order prior and \(\delta_{\mathrm{ord}}\) specifies a tolerance band around the canonical ordering.
We empirically set \(\lambda_{\mathrm{ord}}{=}4\) and \(\delta_{\mathrm{ord}}{=}0.15\).
This term favors assignments that are consistent with the left-to-right FDI ordering and penalizes anatomically implausible permutations, especially in crowded or partially missing dentitions.

We perform one-to-one matching between predicted instances and ground-truth teeth using a Hungarian assignment on the order-aware similarity matrix. In practice, we convert similarities into costs by negation and run the standard (min-cost) Hungarian algorithm~\cite{Hungarian} on the resulting cost matrix.
The complete FDI order-aware matching procedure is summarized in Algorithm~\ref{alg:fhm}.

\begin{algorithm}[t] \caption{FDI Order-Aware Hungarian Matching (FHM)} 
\label{alg:fhm} \small \begin{algorithmic}[1] 
\REQUIRE ~\\
Predicted masks \(\{\mathbf{P}_i\}_{i=1}^{N}\),
class logits \(\{\mathbf{L}_i\}_{i=1}^{N}\),
derived centers \(\{\mathbf{c}_i^{3\mathrm{D}}\}_{i=1}^{N}\);\\
Ground-truth masks \(\{\mathbf{T}_l\}_{l=1}^{L}\), labels \(\{y_l\}_{l=1}^{L}\), ground-truth centers \(\{\hat{\mathbf{c}}_l^{3\mathrm{D}}\}_{l=1}^{L}\).
\ENSURE ~\\
One-to-one assignment \(\pi\) between predictions and ground-truth teeth.
\STATE Compute mask Dice similarities \(S_{\mathrm{mask}}(i,l)\) between \(\sigma(\mathbf{P}_i)\) and \(\mathbf{T}_l\);
\STATE Select class probabilities \(S_{\mathrm{cls}}(i,l) = \sigma(\mathbf{L}_i)[y_l]\);
\STATE Form base similarity \(\mathbf{C}(i,l) = S_{\mathrm{mask}}(i,l)^{\alpha} S_{\mathrm{cls}}(i,l)^{\beta}\);
\STATE Refine \(\mathbf{C}\) with center distance to obtain \(\mathbf{C}_{\mathrm{cent}}\) (Eq.~\eqref{eq:Ccenter});
\STATE Refine \(\mathbf{C}_{\mathrm{cent}}\) with the FDI order prior to obtain \(\mathbf{C}_{\mathrm{ord}}\) (Eq.~\eqref{eq:Cord});
\STATE Convert similarity to cost: \(\mathbf{C}_{\text{cost}} = -\mathbf{C}_{\mathrm{ord}}\); 
\STATE Run the Hungarian algorithm on \(\mathbf{C}_{\text{cost}}\) to obtain the optimal assignment \(\pi\). 
\STATE \textbf{return} \(\pi\) 
\end{algorithmic} \end{algorithm}

Given the assignment \(\pi\), we gather matched predictions \((\mathbf{L}_{\mathrm{match}}, \mathbf{P}_{\mathrm{match}}, \mathbf{s}_{\mathrm{match}}, \mathbf{c}^{3\mathrm{D}}_{\mathrm{match}})\) and the corresponding ground-truth labels, masks, and centers.
We then define the main training objective as
\begin{equation}
\mathcal{L}_{\mathrm{main}}
=
\mathcal{L}_{\mathrm{cls}}
+
2\,\mathcal{L}_{\mathrm{m}}
+
\mathcal{L}_{\mathrm{obj}}
+
\lambda_{\mathrm{cent}}\,\mathcal{L}_{\mathrm{cent}}.
\end{equation}
Here, \(\mathcal{L}_{\mathrm{cls}}\) is a focal classification loss, \(\mathcal{L}_{\mathrm{m}}\) combines Dice and binary cross-entropy losses on the matched masks, \(\mathcal{L}_{\mathrm{obj}}\) is a binary cross-entropy loss that supervises objectness scores with the mask IoU as a soft target, and \(\mathcal{L}_{\mathrm{cent}}\) is a Smooth~L1 loss between the derived 3D centers and their class-mean ground-truth centers (Eq.~\eqref{eq:center_reg}).

To further stabilize the masks, we add a pseudo-mask consistency term \(\mathcal{L}_{\mathrm{pm}}\) with a fixed weight \(w_{\mathrm{pm}}{=}0.2\) and define the full objective as
\begin{equation}
\mathcal{L}
=
\mathcal{L}_{\mathrm{main}}
+
w_{\mathrm{pm}}\,\mathcal{L}_{\mathrm{pm}},
\end{equation}
where \(\mathcal{L}_{\mathrm{pm}}\) denotes the pseudo-mask consistency loss.
Combined with the order-aware similarity-space formulation and center-distance integration in FHM, this objective yields stable, anatomically coherent assignments while retaining the optimality benefits of Hungarian matching.

\section{Experiments}
\label{sec:exp}

\subsection{Experimental Setup}
\label{sec:setup}

\subsubsection{Dataset}
We conduct all experiments on the public 3DTeethSeg'22 benchmark~\cite{3dteethseg}, which contains 1{,}800 intraoral 3D scans (maxilla and mandible) from 900 patients, with 1{,}200 scans for training and 600 for testing according to the official split. 
Raw meshes contain 13k--260k points and typically more than 100k triangles. Following prior work~\cite{TSGC,meshseg}, we downsample each mesh to about 10{,}000 faces using MeshLab~\cite{meshlab} for a uniform resolution.
We formulate 17-class segmentation: 16 tooth classes (central/lateral incisors, canines, first/second premolars, and first/second/third molars on both sides) plus gingiva.
Third molars (wisdom teeth) appear in about 5\% of scans and often co-occur with missing teeth or crowding, making them particularly challenging.
For ablation experiments, we adopt an 80\%/10\%/10\% train/val/test split within the official training set, while comparisons with state-of-the-art methods follow the official training/testing protocol.

\subsubsection{Implementation and Training Details}
The network is implemented in PyTorch and trained on two NVIDIA GeForce RTX~3090 GPUs with a batch size of 4 for 400 epochs. 
We construct $k$-NN graphs with $k{=}32$ and use Adam with an initial learning rate of $\num{1e-3}$. 
To reduce computational overhead, the SAM-guided path is activated after a short warm-up (from epoch 30), and early stopping is applied if the validation loss does not improve for 25 epochs.
For each dental model, the instance decoder outputs $N{=}30$ candidate instances, and masks are binarized with a probability threshold of 0.5.

\subsubsection{Evaluation metrics}
We evaluate using overall accuracy (\(\OA\)) and mean Intersection-over-Union (\(\mIoU\)) over all 17 classes.
Let \(N_c\) be the number of correctly classified faces and \(N_{\mathrm{all}}\) the total number of faces; overall accuracy is
\begin{equation}
\OA \,=\, \frac{N_c}{N_{\mathrm{all}}}.
\end{equation}
Let \(\mathrm{TP}_i\), \(\mathrm{FP}_i\), and \(\mathrm{FN}_i\) denote the true positives, false positives, and false negatives for class \(i\), respectively, and \(K{=}17\) be the number of classes.
The mean IoU is computed as
\begin{equation}
\mIoU \,=\, \frac{1}{K}\sum_{i=1}^{K}\frac{\mathrm{TP}_i}{\mathrm{TP}_i + \mathrm{FP}_i + \mathrm{FN}_i}.
\end{equation}

\subsection{Comparison With State-of-the-Art Methods}
\paragraph{Competing Methods}
We compare SOFTooth against representative baselines on 3DTeethSeg’22~\cite{3dteethseg} and report \(\OA\) and \(\mIoU\):
PointNet~\cite{pointnet}, 
PointNet++~\cite{pointnet++}, 
MeshSegNet~\cite{meshseg}, 
TSGCNet~\cite{TSGC}, 
THISNet~\cite{li2023thisnet}, 
DBGANet~\cite{dbganet}.
For each method, we follow the official configurations or use the authors’ released models and results.

\paragraph{Quantitative Comparison}
Table~\ref{tab:segmentation_performance} reports three-fold cross-validation performance on 3DTeethSeg’22 (17 classes: 16 teeth plus gingiva). 
SOFTooth achieves the best overall accuracy and mean IoU, improving OA from 95.13\% (TSGCNet~\cite{TSGC}) to 96.12\% and mIoU from 84.49\% (THISNet~\cite{li2023thisnet}) to 88.99\%.

Gains are particularly pronounced on minority and morphologically similar classes. 
For third molars (T8/T16), IoU improves from 75.81\%/60.37\% (THISNet~\cite{li2023thisnet}) to 84.67\%/86.22\%, clearly surpassing purely geometric baselines such as TSGCNet~\cite{TSGC} (48.36\%/41.61\%) and DBGANet~\cite{dbganet} (24.15\%/20.46\%). 
On these imbalanced categories, standard training tends to favor frequent teeth and confuse rare posterior classes with neighboring molars~\cite{balancing}.  
By integrating boundary-sensitive 2D--3D fusion, center-consistent masks, and order-aware supervision, SOFTooth better separates tightly arranged teeth and preserves third-molar identities.

For anterior and premolar teeth, SOFTooth also shows consistent gains over existing methods, especially on central and lateral incisors where boundaries are crowded. Although DBGANet~\cite{dbganet} remains competitive on some mid-arch classes, SOFTooth attains the best overall averages by simultaneously improving minority-class recall and boundary precision across both posterior and anterior regions.

\begin{table*}[!t]  
  \centering
  \caption{COMPARISON ON THREE-FOLD CROSS-VALIDATION WITH STATE-OF-THE-ART METHODS}  
  \label{tab:segmentation_performance}  
\adjustbox{max width=\linewidth}{
\begin{tabular}{cccccccccccccccccccc}
\hline
\multirow{2}{*}{Method} & \multicolumn{19}{c}{Metric} \\ \cline{2-20} 
 & OA & mIoU & BG & T1 & T2 & T3 & T4 & T5 & T6 & T7 & T8 & T9 & T10 & T11 & T12 & T13 & T14 & T15 & T16 \\ \hline
PointNet~\cite{pointnet} & 82.85 & 54.91 &84.07& 63.35 & 55.77 & 56.46 &60.19  & 54.52 & 61.52& 49.74 &19.56  & 63.23 &57.14  &58.21  &61.00  &54.15  &61.51 &52.51  &20.60   \\
PointNet++~\cite{pointnet++} & 88.96  &68.54 & 88.28 &	70.57 &	68.56& 	70.93& 	73.92 &	69.30 &	74.79 	&67.66 &	43.88 &	71.01 &	68.47 &	70.90 &	74.95 &	70.74 &	76.22 &	69.66& 	35.42 \\
MeshSegNet~\cite{meshseg} & 95.03&	82.96&	95.86 	&89.84& 	88.96 &	88.40 &	88.23 &	82.02 	&85.46 &	80.31 &	73.02 &	88.92 &	87.95 &	89.50 	&88.27 &	82.02& 	83.69 &	73.67 &	44.27 \\
TSGCNet~\cite{TSGC} & 95.24 & 81.82 & \textbf{96.59} & 92.20 & 91.17 & 90.07 & 87.91 & 80.98 &83.87 & 77.09 & 48.36 & 91.88 & 90.58 & 90.93 & 88.91 & 80.74 & 82.40 & 75.75 & 41.61 \\
THISNet~\cite{li2023thisnet} & 94.46&	84.29&	93.60 &	92.73 &	91.49 &	90.06& 	88.95 &	79.79 &	84.06 &	76.76 &	75.81 &	92.04 &	90.81 	&91.20 &	89.13 	&81.72 &	82.46 &	72.10 &	60.37  \\
DBGANet~\cite{dbganet} & 93.21 & 77.46 & 93.71 & 84.69 & 83.93 & 85.40 & 87.14 & 83.94 & 84.73 &77.85 & 24.15 & 84.82&84.49 & 86.04 & 87.47 & \textbf{85.82} & 84.96 & 77.27 & 20.46 \\
\textbf{Ours} & \textbf{96.12} & \textbf{88.99} & 96.31 & \textbf{93.59} & \textbf{92.35} & \textbf{90.70} & \textbf{90.44} & \textbf{85.59} & \textbf{86.64} & \textbf{81.37} & \textbf{84.67} & \textbf{93.69} & \textbf{92.39} & \textbf{92.13} & \textbf{91.29} & 85.45 & \textbf{87.24} & \textbf{82.75} & \textbf{86.22} \\ \hline

\end{tabular}
}
\end{table*}

\begin{figure*}[!t]
    \centering
    \includegraphics[width=1\linewidth]{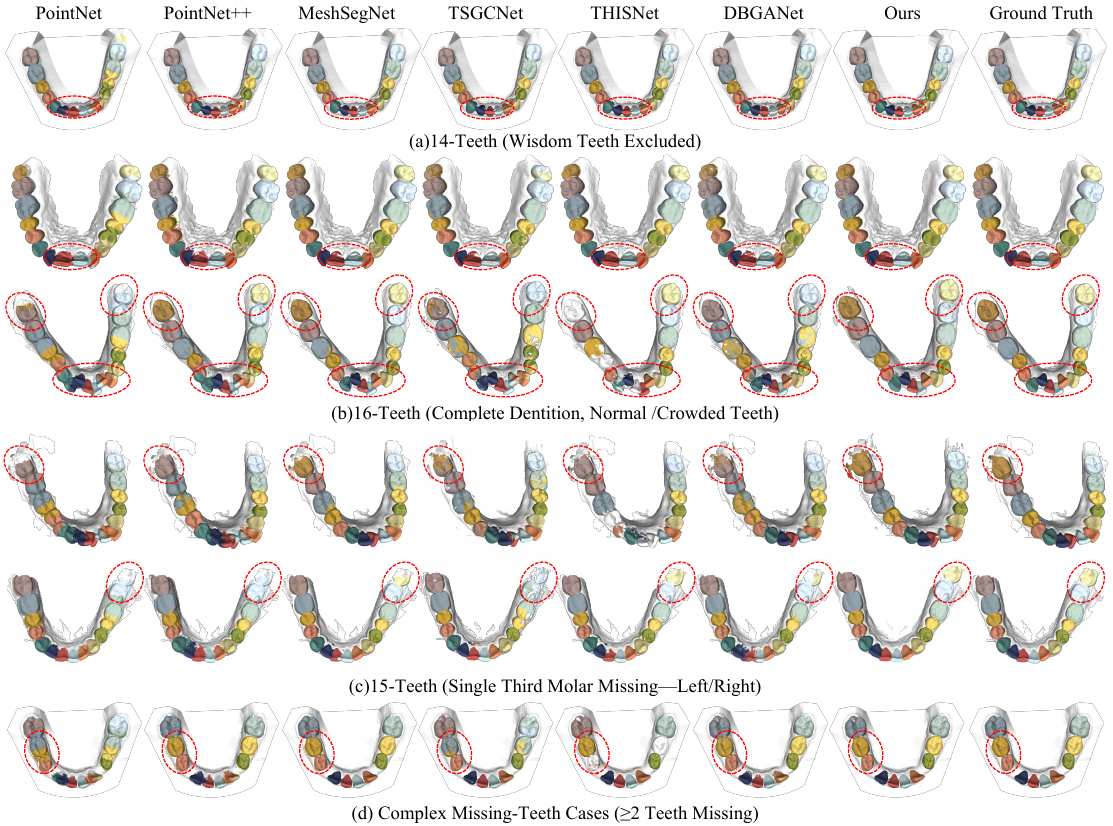}
    \caption{Qualitative results compared with state-of-the-art methods. (a) 14-Teeth (wisdom teeth excluded), (b) 16-Teeth (complete dentition; normal/crowded), (c) 15-Teeth (single third molar missing—left/right), and (d) complex partial dentitions ($ \ge $2 missing teeth). Red dotted circles indicate segmentation distinctions of recent state-of-the-art methods, and the proposed method maintains better segmentation performance and robustness.}
    \label{fig:placeholder2}
\end{figure*}
\paragraph{Qualitative Results}
Fig.~\ref{fig:placeholder2} visualizes qualitative comparisons with representative methods across four scenarios: (a) 14 teeth (third molars excluded), (b) 16 teeth (complete dentition; normal or crowded), (c) 15 teeth (single third molar missing), and (d) complex partial dentitions (at least two missing teeth).

In scenario (a), all methods perform well on complete arches without third molars, but our SOFTooth produces cleaner inter-tooth boundaries than PointNet~\cite{pointnet} and PointNet++~\cite{pointnet++}, which occasionally confuse adjacent teeth and bleed into the gingiva due to limited local aggregation and boundary modeling. 
In well-aligned 16-tooth arches (scenario (b)), most methods (except PointNet) produce accurate masks, whereas in crowded arches, PointNet, TSGCNet, THISNet, and DBGANet~\cite{pointnet,TSGC,li2023thisnet,dbganet} frequently misclassify third molars as second molars, exhibiting boundary leakage and center drift. THISNet further filters candidate instances using a fixed confidence threshold in the assignment stage, so low-confidence tooth instances, especially rare or partially erupted teeth such as third molars, are often discarded and misclassified as gingiva. PointNet++ and MeshSegNet~\cite{pointnet++,meshseg} generally predict the correct number of teeth but still show boundary sticking in crowded regions. In contrast, SOFTooth assigns instances based jointly on mask similarity, class probability, center distance, and anatomical tooth order instead of a hard score cutoff, yielding sharper tooth-gingiva and inter-tooth boundaries.

In scenarios (c) and (d), which involve various missing-tooth patterns, several baselines (e.g., PointNet, PointNet++, TSGCNet, and THISNet) tend to classify the last visible tooth as a second molar, producing duplicated second-molar labels and sometimes propagating this shift to more anterior teeth or violating the anatomical ordering along the arch. In contrast, SOFTooth employs FHM, which encourages predictions that respect the realistic spatial layout of teeth, yielding coherent FDI indices and reliable instance masks even in cases with a single third molar, partially erupted teeth, or multiple missing teeth. These qualitative trends are consistent with the quantitative gains observed for posterior teeth, particularly third molars.

\subsection{Ablation Studies}
To validate the effectiveness of the proposed components, we perform ablation experiments on the 3DTeethSeg'22 dataset. 
Unless otherwise specified, we report Overall Accuracy (OA), mean Intersection-over-Union (mIoU), and the IoU of third molars to assess both global and minority-class performance. Results are summarized in Table \ref{tab:ABLATION}, with detailed analysis below. Among these settings, the first setting is the baseline network, which contains a two-stream geometric encoder, an instance decoder, and vanilla Hungarian matching based on mask--class similarity only, without PRG, CMR, or FHM. The loss contains only the focal classification term, the mask loss, and the objectness loss, without the center regularizer or pseudo-mask consistency.

\begin{table}[h]
\centering
\caption{ABLATION RESULTS FOR KEY COMPONENTS: PRG, CMR, AND FHM.}  
\label{tab:ABLATION}
\resizebox{\linewidth}{!}{ 
\begin{tabular}{ccccccc}
\hline
\multirow{2}{*}{No.} & \multicolumn{3}{c}{Settings} & \multirow{2}{*}{OA} & \multirow{2}{*}{mIoU} & \multirow{2}{*}{3rd Molars IoU} \\ \cline{2-4}
                     & PRG    & CMR    & FHM    &                         &                           &                                                   \\ \hline
1          &         &          &         & 94.76                   & 84.49                     & 69.45                                             \\ 
2                    & \checkmark &          &         & 95.41                   & 87.19                     & 75.90                                             \\ 
3                    &         & \checkmark        &         & 95.48                   & 84.75                     & 67.29                                             \\ 
4                    &         &          & \checkmark       & 95.92                   & 86.09                     & 69.61                                             \\ 
5                    & \checkmark  & \checkmark        &         & 95.89                   & 86.43                    & 68.69                                             \\ 
6                    & \checkmark   &          & \checkmark       & 96.00                   & 87.70                     & 77.27                                             \\ 
7                    &         & \checkmark        & \checkmark       & 96.00                   & 87.67                     & 77.11                                             \\ 
8(Ours)              & \checkmark  & \checkmark        & \checkmark       & \textbf{96.12}          & \textbf{88.99}            & \textbf{85.45}                                    \\ \hline
\end{tabular}
}
\end{table}

\subsubsection{Effectiveness of PRG}
\begin{figure*}
    \centering
    \includegraphics[width=\linewidth]{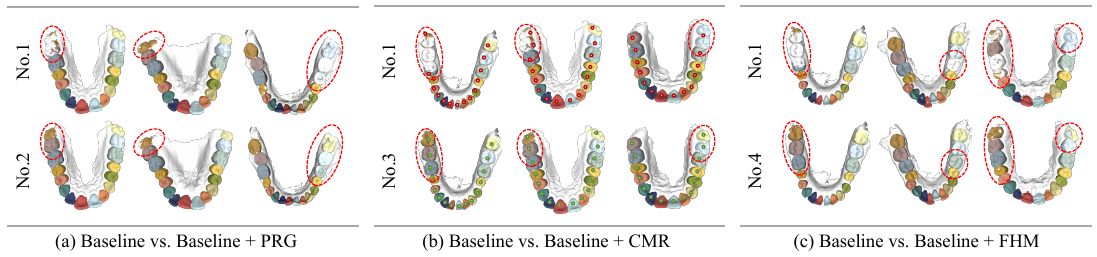}
    \caption{Qualitative ablation of the proposed modules on 3D tooth instance segmentation (corresponding to settings No.~1--No.~4 in Table~\ref{tab:ABLATION}). Red dotted circles highlight regions where the baseline and our PRG/CMR/FHM variants differ in tooth-gingiva leakage, inter-tooth boundary delineation, center drift, and FDI label consistency.
  (a) Baseline (No.~1) vs.\ Baseline + PRG (No.~2).
  (b) Baseline (No.~1) vs.\ Baseline + CMR (No.~3).
  (c) Baseline (No.~1) vs.\ Baseline + FHM (No.~4).}
    \label{fig:ablation_visualization}
\end{figure*}

Table~\ref{tab:ABLATION} (No.~1 vs.\ No.~2) shows that adding PRG to the baseline improves OA from 94.76\% to 95.41\%, mIoU from 84.49\% to 87.19\%, and third-molar IoU from 69.45\% to 75.90\%.
These gains indicate that injecting SAM-derived occlusal-view semantics as boundary-sensitive priors effectively enhances 3D tooth instance segmentation.
Qualitatively, Fig.~\ref{fig:ablation_visualization}(a) shows that PRG reduces tooth-gingiva leakage and better separates adjacent crowns, especially around second and third molars. Without PRG, the baseline tends to blur contacts and sometimes leaks into gingiva in crowded arches. With PRG, geometry-aligned projection and center-guided gating produce sharper tooth contours and more coherent tooth shapes.

We further evaluate pairwise confusion rates between anatomically adjacent tooth types, defined as the percentage of teeth in a given pair that are assigned to the wrong class within that pair.
As shown in Table~\ref{tab:pair_confusion}, removing PRG increases confusion between second and third molars (pair\_2nd3rd), premolars and first molars (pair\_pre\_mo), and central and lateral incisors (pair\_cen\_lat), especially on scans containing third molars, indicating that PRG improves the separability of neighboring instances by sharpening boundaries and reinforcing tooth-specific semantics.
\begin{table}[t]
\centering
\caption{Pairwise confusion rates (\%, $\downarrow$) between anatomically adjacent tooth types: (i) second vs.\ third molars (pair\_2nd3rd), (ii) premolars vs.\ first molars (pair\_pre\_mo), and
(iii) central vs.\ lateral incisors (pair\_cen\_lat) on the full test set and the subset containing third molars.}
\label{tab:pair_confusion}
\resizebox{\linewidth}{!}{
\begin{tabular}{llccc}
\hline
Setting & Split & pair\_2nd3rd $\downarrow$ & pair\_pre\_mo $\downarrow$ & pair\_cen\_lat $\downarrow$ \\
\hline
w/o PRG & Full set      & 0.80 & 1.26 & 0.69 \\
w/o FHM & Full set      & 0.79 & 1.34 & 0.98 \\
w/ PRG+FHM  & Full set      & \textbf{0.66} & \textbf{0.79} & \textbf{0.57} \\
\hline
w/o PRG & Subset   & 3.53 & 1.51 & 4.15 \\
w/o FHM & Subset   & 3.05 & 1.81 & 7.03 \\
w/ PRG+FHM  & Subset   & \textbf{2.46} & \textbf{1.50} & \textbf{2.18} \\
\hline

\end{tabular}
}
\end{table}
\subsubsection{Effectiveness of CMR}
We next analyze the impact of CMR on center localization.
Following Sec.~\ref{sec:method}, we derive a 3D center for each predicted instance from its mask and measure the normalized center error as the L1 distance between the predicted and class-mean ground-truth centers, divided by the diagonal length of the dental model bounding box.
As reported in Table~\ref{tab:ABLATION}, introducing CMR (No.~3) yields a small overall improvement in OA (from 94.76\% to 95.48\%) and mIoU (from 84.49\% to 84.75\%), while providing more stable center localization.
Table~\ref{tab:center_error} shows that CMR reduces the center error from 0.61\% to 0.58\% on the full test set and from 1.08\% to 1.01\% on the third-molar subset. Although the absolute improvements are small, they are systematic and indicate more accurate instance localization.

Qualitative results in Fig.~\ref{fig:ablation_visualization}(b) show that CMR corrects center drift in crowded or partially erupted regions.
By linking masks to geometric centroids and imposing class-consistent pseudo-mask supervision, CMR reduces fragmented or shifted instances and yields more reliable tooth centers, which in turn benefit the subsequent FHM-based assignment.
\begin{table}[t]
\centering
\caption{Average normalized center error (\%, $\downarrow$) with and without CMR on the full test set and on the subset containing third molars.}
\label{tab:center_error}
\resizebox{0.6\linewidth}{!}{
\begin{tabular}{lcc}
\hline
Setting      & Full set $\downarrow$ & Subset $\downarrow$ \\
\hline
w/o CMR      & 0.61 & 1.08 \\
w/ CMR       & \textbf{0.58} & \textbf{1.01} \\
\hline
\end{tabular}
}
\end{table}

\subsubsection{Effectiveness of FHM}
FHM primarily targets anatomical consistency and label reliability by incorporating FDI tooth sequence and center distance into the matching cost. As shown in Table~\ref{tab:ABLATION} (No.~4 vs.\ No.~1), adding FHM on top of the baseline improves OA from 94.76\% to 95.92\% and mIoU from 84.49\% to 86.09\%, while keeping the third-molar IoU at a similar level (69.45\% vs.\ 69.61\%), indicating that FHM mainly enhances global assignment quality rather than directly boosting minority-class segmentation.
Table~\ref{tab:pair_confusion} further shows that FHM reduces pairwise confusion between anatomically adjacent teeth on both the full test set and the third-molar subset. Compared with the variant without PRG, which exhibits higher confusion on the third-molar subset, the full SOFTooth model with PRG and FHM attains the lowest pairwise confusion, indicating that order-aware assignment effectively complements boundary-sensitive fusion.

Qualitative results in Fig.~\ref{fig:ablation_visualization}(c) are consistent with these trends: FHM suppresses anatomically implausible label shifts (e.g., duplicated second-molar labels when third molars are present) and yields FDI indices that better respect the spatial layout along the arch, particularly in cases with missing or crowded teeth.

\subsubsection{Synergistic Effects of Component Combinations}
Table~\ref{tab:ABLATION} further evaluates pairwise combinations (No.~5–7) and the full model (No.~8). Compared with single-component variants (No.~2–4), combining PRG with CMR (No.~5) or FHM (No.~6) yields consistent gains in OA and mIoU, and noticeably improves third-molar IoU, indicating that boundary-sensitive 2D--3D fusion is complementary to both center-consistent refinement and order-aware assignment. The combination of CMR and FHM (No.~7) mainly stabilizes instance localization and promotes anatomically coherent labeling, but its improvement on third-molar IoU is relatively modest, underscoring the role of PRG in strengthening minority posterior teeth. The full SOFTooth model (No.~8), which integrates PRG, CMR, and FHM, achieves the best overall performance across OA, mIoU, and third-molar IoU, demonstrating that boundary-sensitive fusion, center-guided refinement, and FDI-guided matching are jointly necessary for robust 3D tooth instance segmentation in cases with missing teeth, crowding, and third molars.


\subsection{Discussion}
\label{sec:summary}
Our experiments show that the three proposed components are complementary and jointly enable robust tooth instance segmentation. FHM augments the similarity space with FDI tooth order and 3D center distance, reducing mis-assignments in crowded arches and complex partial dentitions, as reflected by lower pairwise confusion between neighboring teeth and more consistent tooth indices along the arch. CMR links instance masks to geometric centroids via pseudo-mask supervision and a center-based regularizer, stabilizing instance localization and alleviating mask--center drift. Built on this geometrically consistent representation, PRG injects frozen SAM occlusal-view embeddings as boundary-sensitive priors into native 3D features, sharpening tooth-gingiva and inter-tooth boundaries and markedly improving the IoU of minority posterior teeth (e.g., third molars) without any tooth-specific 2D annotations. The full model, which combines PRG, CMR, and FHM, therefore benefits simultaneously from strong local semantics, reliable instance centers, and anatomically coherent supervision while remaining comparable to native 3D methods in supervision type and training cost. Future work will focus on validating SOFTooth on larger multi-center IOS datasets and exploring more flexible, learned anatomical priors to further improve assignment robustness under highly irregular dentitions and severe malocclusions.

\section{Conclusions}
We proposed SOFTooth, a unified 2D--3D framework for 3D tooth instance segmentation in challenging clinical conditions with crowding, missing teeth, and rare third molars. The Point-wise Residual Gating module injects boundary-sensitive occlusal-view semantics from a frozen SAM encoder into native 3D features, the Center-Guided Mask Refinement module enforces center-mask geometric consistency to stabilize instance localization, and the FDI Order-Aware Hungarian Matching module incorporates anatomical tooth order and 3D center distance into the assignment stage to preserve coherent FDI indices.
On the 3DTeethSeg'22 benchmark, SOFTooth achieves state-of-the-art overall accuracy and mean IoU, with particularly strong gains on posterior teeth and third molars. Auxiliary analyses show reduced confusion between adjacent teeth and lower 3D center error, and qualitative results confirm more reliable segmentation in crowded arches and partial dentitions. The framework requires only a single occlusal-view SAM pass without any tooth-specific 2D labels and remains compatible with standard 3D backbones, making it practical for integration into real-world digital dentistry pipelines.

\bibliographystyle{IEEEtran}  
\bibliography{main}
\end{document}